\def\paperID{0000}
\def\confName{CVPR}
\def\confYear{2024}

\def\authorBlock{
Chao He,~~Jianqiang Ren,~~Yuan Dong,\\~~Jianjing Xiang,~~Xiejie Shen,~~Weihao Yuan,~~Liefeng Bo \\ 

   Tongyi Lab,~~Alibaba Group\\
   
 {\tt\small \{yichao.hc, jianqiang.rjq, dy283090, jianjing.xjj, }\\
 {\tt\small \ shenxiejie.sxj, qianmu.ywh, liefeng.bo\}@alibaba-inc.com}
}

\newif\ifreview 
\newif\ifarxiv \newcommand{\arxiv}{\arxivtrue}
\newif\ifcamera 
\newif\ifrebuttal 

\arxiv  

\pdfoutput=1
\documentclass[10pt,onecolumn,letterpaper]{article}
\ifreview \usepackage[review]{cvpr} \fi
\ifarxiv \usepackage[pagenumbers]{cvpr} \fi
\ifrebuttal \usepackage[rebuttal]{cvpr} \fi
\ifcamera \usepackage{cvpr} \fi

\usepackage{graphicx}	
\usepackage{amsmath}	
\usepackage{amssymb}	
\usepackage{booktabs}
\usepackage{times}
\usepackage{microtype}
\usepackage{epsfig}
\usepackage[table,xcdraw,dvipsnames]{xcolor}
\usepackage{caption}
\usepackage{float}
\usepackage{placeins}
\usepackage{color, colortbl}
\usepackage{stfloats}
\usepackage{enumitem}
\usepackage{tabularx}
\usepackage{xstring}
\usepackage{multirow}
\usepackage{xspace}
\usepackage{url}
\usepackage{subcaption}
\usepackage{xcolor}
\usepackage[hang,flushmargin]{footmisc}

\ifcamera \usepackage[accsupp]{axessibility} \fi

% ...

\ifarxiv  \fi

\newcommand{\R}[1]{{%
    \textbf{%
        \ifstrequal{#1}{1}{\textcolor{red}{R#1}}{%
        \ifstrequal{#1}{2}{\textcolor{blue}{R#1}}{%
        \ifstrequal{#1}{3}{\textcolor{magenta}{R#1}}{%
        \ifstrequal{#1}{4}{\textcolor{teal}{R#1}}{%
                           \textcolor{cyan}{R#1}%
        }}}}%
    }%
}}

\usepackage{xr-hyper}

\makeatletter
\newcommand*{\addFileDependency}[1]{
  \typeout{(#1)}
  \@addtofilelist{#1}
  \IfFileExists{#1}{}{\typeout{No file #1.}}
}

\makeatother
\newcommand*{\myexternaldocument}[1]{
    \externaldocument{#1}
    \addFileDependency{#1.tex}
    \addFileDependency{#1.aux}
}

\definecolor{cvprblue}{rgb}{0.21,0.49,0.74}
\usepackage[pagebackref,breaklinks,colorlinks,citecolor=cvprblue]{hyperref}
\usepackage[capitalize]{cleveref}
\crefname{section}{Sec.}{Secs.}
\crefname{table}{Table}{Tables}
\crefname{figure}{Fig.}{Figs.}

\ifarxiv \crefname{appendix}{App.}{Apps.}
\else \crefname{appendix}{Suppl.}{Suppls.} \fi

\usepackage{indentfirst}
\unless\ifarxiv \myexternaldocument{_supplementary} \fi

\begin{document}
 
\title{Textoon: Generating Vivid 2D Cartoon Characters from Text Descriptions}

\author{\authorBlock}
\maketitle

\begin{figure*}[h]
\vspace{-20pt}
\centering
\includegraphics[width=0.98\textwidth]{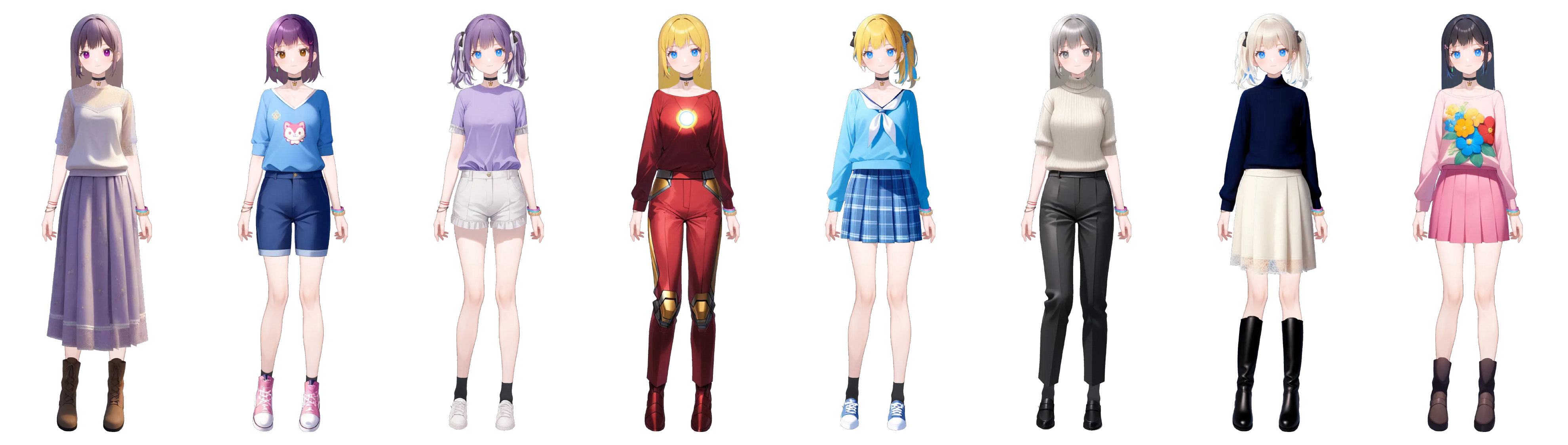}
\caption{Examples of animatable 2D cartoon characters generated by Textoon.}
\label{fig:live2d-results}
 
\end{figure*}

\begin{abstract}


The 2D cartoon style is a prominent art form in digital character creation, particularly popular among younger audiences. While advancements in digital human technology have spurred extensive research into photorealistic digital humans and 3D characters, interactive 2D cartoon characters have received comparatively less attention. Unlike 3D counterparts, which require sophisticated construction and resource-intensive rendering, Live2D, a widely-used format for 2D cartoon characters, offers a more efficient alternative, which allows to animate 2D characters in a manner that simulates 3D movement without the necessity of building a complete 3D model. Furthermore, Live2D employs lightweight HTML5 (H5) rendering, improving both accessibility and efficiency. In this technical report, we introduce Textoon, an innovative method for generating diverse 2D cartoon characters in the Live2D format based on text descriptions. The Textoon leverages cutting-edge language and vision models to comprehend textual intentions and generate 2D appearance, capable of creating a wide variety of stunning and interactive 2D characters within one minute. The project homepage
is \url{https://human3daigc.github.io/Textoon_webpage/}.

\end{abstract}

\section{Introduction}
\label{sec:intro}


Cartoon characters are fictional beings often distinguished by their adorable appearances and vivid colors. They are widely used in films, games, social media, and advertising. Cartoon characters can be crafted in both 2D and 3D styles. While 3D animations provide greater freedom for creation and control, they typically come with higher costs and depend on rendering engines. Conversely, creating 2D cartoon digital characters is more straightforward and efficient, making them especially suitable for devices with limited processing power, such as mobile phones and web applications. In the realm of 2D cartoon characters, Live2D\cite{live2d} has emerged as a leading standard for delivering real-time interactive performance.

Live2D is a technology used to create 2D character models with 3D-like interactivity.
Live2D works by using a model that consists of a base illustration and a set of control points that define how different parts of the character can move. These movements can be controlled, allowing for a wide range of expressions and actions. Live2D effectively converts original 2D artwork into dynamic, animated characters. Its user-friendly, 2D interface makes it accessible for new illustrators and designers. Additionally, the lightweight models are highly compatible across various platforms, including HTML5 (H5).

While Live2D has made it easier to create animated 2D characters, the detailed processes of layering and mesh binding can still be time-consuming and labor-intensive. Additionally, modifying existing Live2D models to achieve different appearances remains a challenge. To tackle these issues, we present Textoon—a framework designed to generate diverse Live2D models from text descriptions. Utilizing existing Live2D models and advanced generative technologies, Textoon empowers users to create customized Live2D models with straightforward textual prompts. The key features of Textoon include:

\noindent {\bf Accurate Text Parsing.}
Our text parsing model excels at extracting detailed information from complex user descriptions. It accurately identifies features such as back hair, side hair, bangs, eye color, eyebrows, face shape, clothing type, and shoe type. This advanced text parsing capability allows for more flexible user inputs. 

\noindent {\bf Controllable Appearance Generation.}
After parsing the text, each component is synthesized into a comprehensive character template. The contour boundaries offer precise control over the shape of the generated character, while a text-to-image model takes charge of generating the inner color and texture.

\noindent {\bf Editable.}
If users are not satisfied with the initial generated result and wish to modify specific details,our framework provides assistance in selecting specific positions to add, remove, or modify elements.


\noindent {\bf Animation.}
The control coefficients for the Live2D model's mouth primarily include MouthOpenY and MouthForm. MouthOpenY controls the vertical movement of the mouth, while MouthForm adjusts the expressions, such as upturning and grimacing. However, these controls often result in suboptimal driving performance. To enhance the accuracy of speech animations for cartoon characters, we integrate ARKit's face blend shape capabilities into the Live2D lip-sync functionality. This integration significantly improves the realism and precision of the animated speech.

Our main contributions in this work are as follows:

\begin{itemize}
  \item To the best of our knowledge, Textoon is the first method that enables the generation of Live2D characters from text prompts, capable of creating a new 2D cartoon character within a minute without the need for manual binding.

  \item We fine-tune the LLM to accurately extract descriptive terms for each body component from complex texts, ensuring that the generated results closely align with the user's input. Additionally, we leverage the capabilities of text-to-image models to create a wide variety of stunning animatable 2D cartoon characters.

  \item  By refactoring the Live2D facial animation mechanism, we significantly improve facial expressiveness.
\end{itemize}

\section{Related Work}
\label{sec:related}
\noindent {\bf Text-to-digital human.} ChatAvatar\cite{10.1145/3588430.3597244} utilizes text inputs to generate facial assets with ultra-high-resolution textures, revolutionizing the traditional process of creating 3D assets. Using a diffusion model and a comprehensive facial asset dataset, it produces computer-generated (CG) assets that are compatible with mainstream rendering engines through text-based interactions, thereby simplifying the user experience. However, the generated outputs are incomplete and lack elements such as hair, eyeballs, bodies, and clothing, making them difficult for users to use directly.

\begin{figure*}[tb]
\centering
\includegraphics[width=0.98\textwidth]{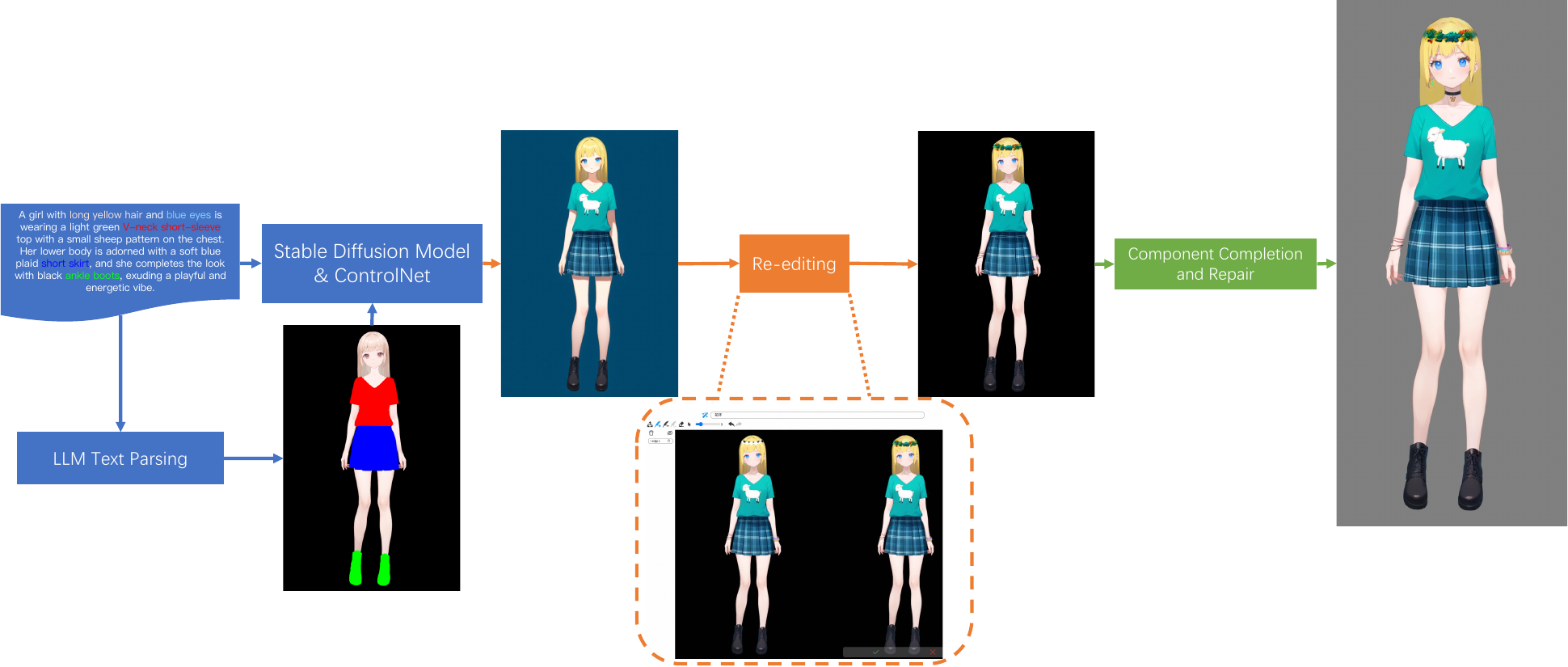}
\caption{Pipeline of the Textoon. The framework includes text parsing, controllable appearance generation, re-editing, and component completion and repair modules.}
\label{fig:pipeline-text2live2d}
\vspace{-5pt}
\end{figure*}

Make-A-Character\cite{ren2023makeacharacter} allows users to create high-quality, fully detailed, and animatable 3D digital humans through simple text descriptions. Users can specify features such as facial shape, eye characteristics, iris color, hairstyles and their colors, types of eyebrows, mouths, and noses, as well as the addition of wrinkles and freckles. However, the generated results require a powerful rendering engine and incur high computational costs to render.


\noindent {\bf Diffusion models and ControlNet.} Stable Diffusion (SD)\cite{rombach2021highresolution} exemplifies the powerful capabilities of diffusion models by integrating the UNet framework to iteratively generate images conditioned on text descriptions. ControlNet\cite{zhang2023adding} enhances the control and flexibility of generative models, including diffusion models. It introduces additional control signals or constraints into the generation process, thereby improving the accuracy and consistency of the generated results. This approach significantly expands the range of applications and use cases of generative models. Through conditional generation methods, ControlNet can produce high-quality data that meet specific constraints or requirements, showcasing powerful functionality and potential in areas such as artistic creation, virtual reality, and film production.

\noindent {\bf Large Language Models.} Large Language Models (LLMs) typically refer to language models with hundreds of billions or even more parameters. Using vast amounts of data, powerful computational hardware, and Transformer architectures\cite{vaswani2017attention}, these language models have been scaled to unprecedented sizes. Early research on LLM, such as T5\cite{raffel2020exploring}, employed transfer learning techniques. Subsequently, GPT-3\cite{brown2020language} demonstrated that LLMs could achieve zero-shot transfer to downstream tasks without the need for fine-tuning. When provided with task descriptions and example prompts, LLMs can generate accurate responses. For example, the latest Qwen2.5 model\cite{yang2024qwen2}, trained on an exceptionally large dataset of 18 trillion tokens, excels at following instructions, generating extended texts, understanding structured data and producing structured outputs.

\section{Live2D Generation}

This section provides a detailed introduction to our approach. We begin with a brief review of the implementation principles of Live2D and introduce our component splitting method designed to enhance generation diversity. Then, we present our carefully designed modules for text parsing, controllable appearance generation, re-editing, and component completion, which enable us to generate a brand-new Live2D character from a single sentence within a minute, as shown in Fig.~\ref{fig:pipeline-text2live2d}.

\subsection{Preliminary of Live2D}
\begin{figure}[tb]
    \centering
    \includegraphics[width=0.4\linewidth]{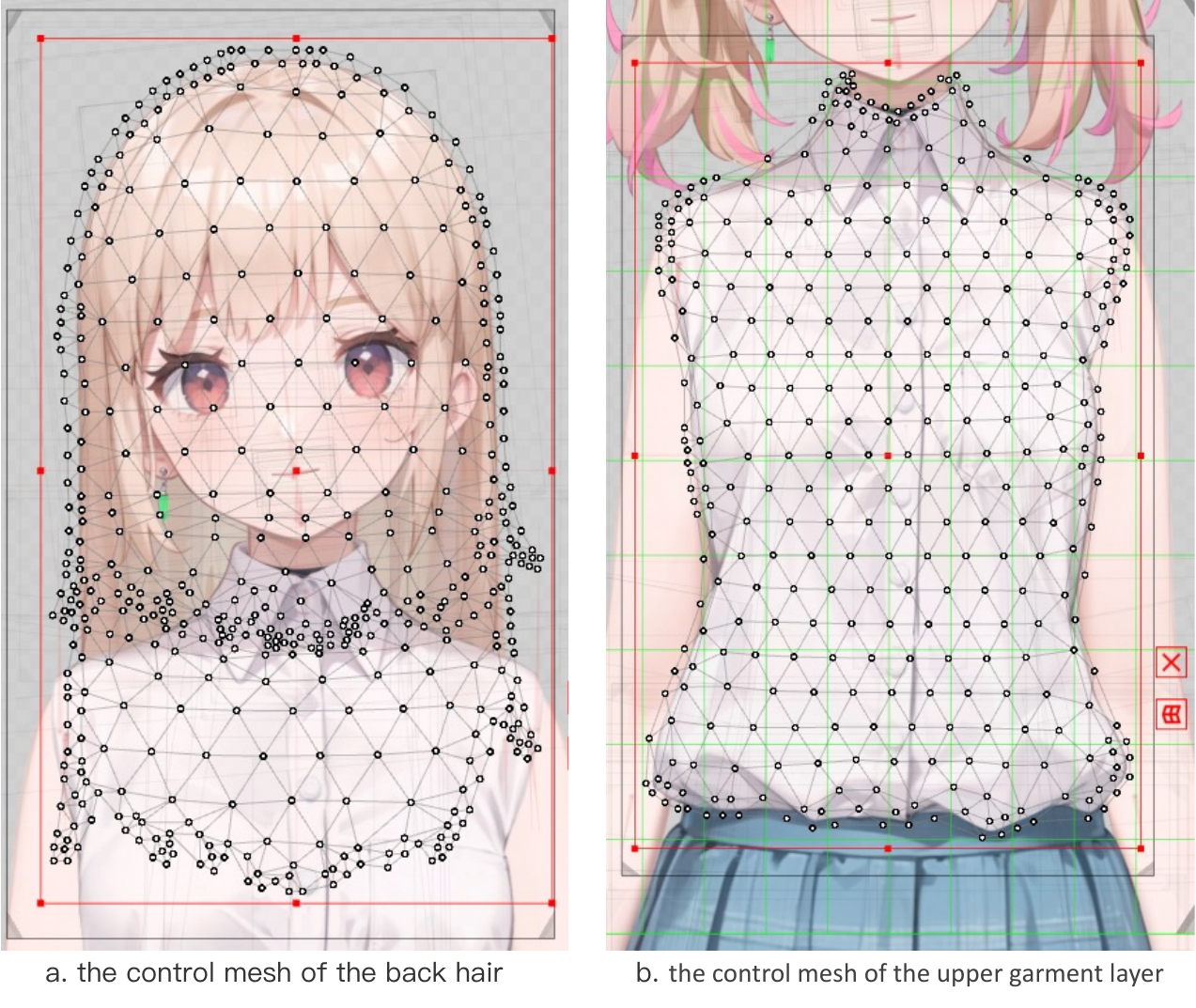}
    \caption{Meshes of different layers.}
    \vspace{-10pt}
    \label{fig:live-mesh}
\end{figure}

A typical Live2D character comprises multiple layers, including the body, clothing, hair, and other elements. Each layer is segmented into a polygon mesh, which controls the respective parts of the 2D character, as illustrated in Fig.~\ref{fig:live-mesh}. This layered composition facilitates smooth deformations by manipulating the positions of the mesh points.

\subsection{Component Splitting}

\begin{figure}[tb]
    \centering
    \includegraphics[width=0.4\linewidth]{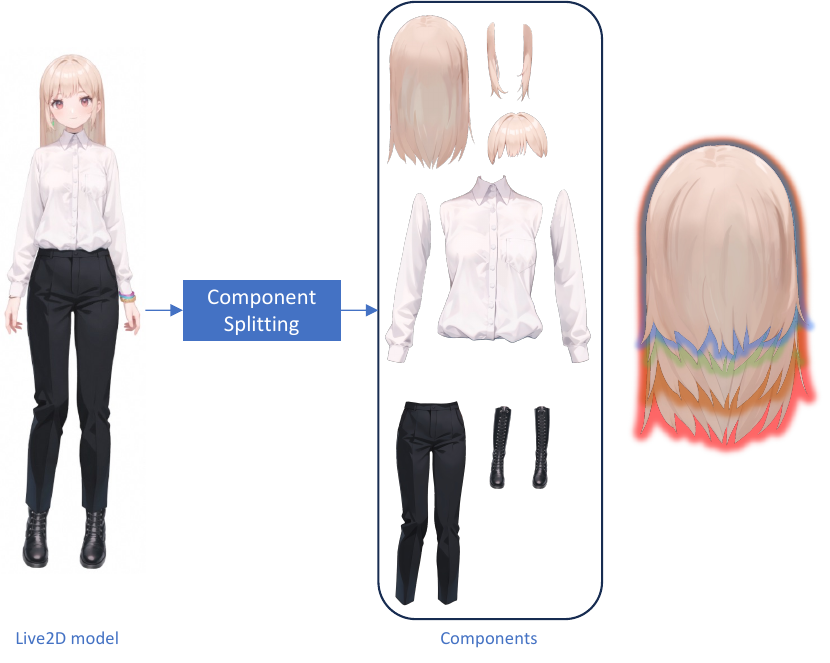}
    \caption{Splitting model components, larger elements can be utilized to create short variations.}
 
    \label{fig:component-splitting}
\end{figure}

Given that Live2D models consist of numerous layers, we decide to merge some intricate, smaller layers to decrease the overall layer count. While this may slightly impact the expressiveness of detailed movements, it simplifies the generation process. For each body part, we choose to use larger elements to generate smaller-scale elements, thereby enhancing the models' diversity. As illustrated in Fig.~\ref{fig:component-splitting}, the long hair element can be utilized to create short hair variations.

\subsection{Text Parsing}
\begin{figure}[tb]
    \centering
    \includegraphics[width=0.6\linewidth]{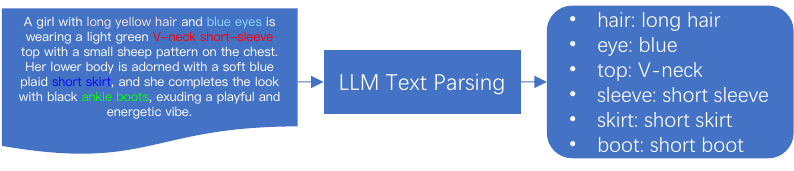}
    \caption{Using the fine-tuned LLM to parse component categories from complex input text.}
    \vspace{-10pt}
    \label{fig:text-parsing}
\end{figure}

We need to parse the corresponding body parts from the text input and then combine them to control the next step of generation. Large language models are unable to directly extract suitable component words from complex and varied input texts. To address this, we generate descriptive text by randomly combining existing components with commonly used words, resulting in 640,000 text-component pairs. This generated data is used to fine-tune the Qwen2.5-1.5B\cite{yang2024qwen2} model. As shown in Fig.~\ref{fig:text-parsing}, we can accurately parse component categories from complex input text at millisecond speeds within 4GB of memory (RTX 4090), achieving an accuracy rate of more than 90\%.

\subsection{Controllable Appearance Generation}
The facial proportions of 2D cartoon characters tend to remain consistent, with their unique traits primarily found in their hair, clothing, and accessories. Furthermore, the high resolution of 2D original artwork makes the choice of image generation model especially important.

We evaluated the top text-to-image models based on control accuracy, generation quality, text relevance, and their ability to create text or patterns. Ultimately, we selected SDXL \cite{podell2023sdxl} as the optimal choice. SDXL excels in controllability, produces images with sharp edges, and supports a maximum resolution of 1024 pixels. Additionally, it handles long text descriptions effectively and generates precise text patterns accurately.

\begin{figure}[tb]
    \centering
    \includegraphics[width=0.4\linewidth]{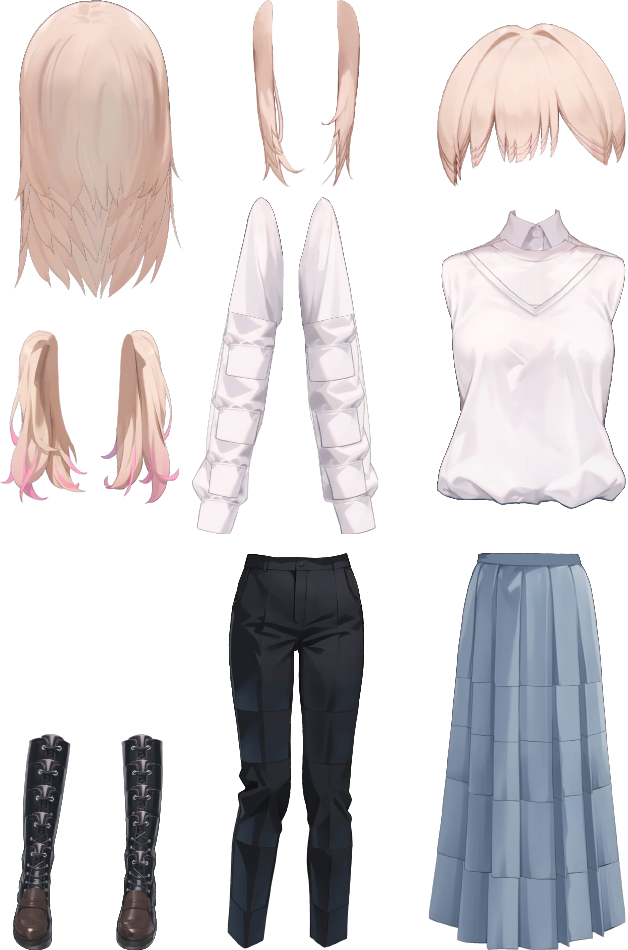} 
    \caption{The divisions of each component within our template model.}
 
    \label{fig:part-list}
\end{figure}

To maintain the model's driving performance, it is essential that the generated output adhere to the specified areas for each component. In our template model, we categorize the components as follows: 5 types of back hair, 3 types of mid hair, 3 types of front hair, 5 types of tops, 6 types of sleeves, 5 types of pants, 5 types of skirts, and 6 types of shoes, as shown in Fig.~\ref{fig:part-list}. By combining these components and utilizing the control features of the base model, we have managed to achieve a wide variety of outputs while preserving the original driving performance.

\subsection{Re-editing}
We employ an image editing technique\cite{liu2024magicquill} that enables users to freely sketch specific areas and annotate these sketches with text to refine the details of the initial character image. Once the user's specifications are satisfied, the adjusted character image is finalized as the final character appearance.

\subsection{Component Completion}
\begin{figure}[tb]
    \centering
    \includegraphics[width=1\linewidth]{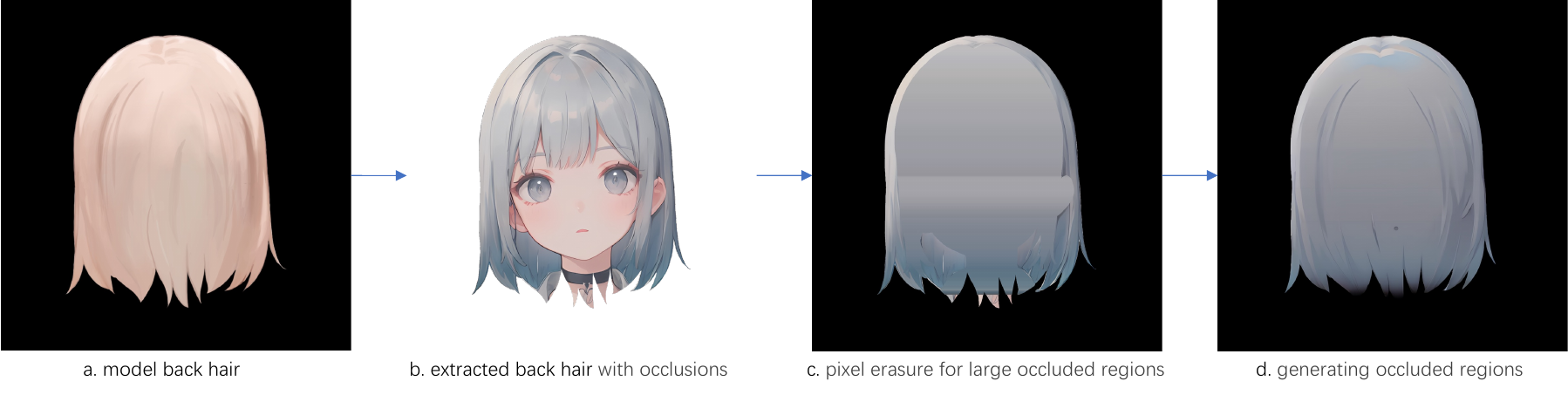}
    \caption{Restoring the back hair: First, extract the pixels (b) from the generated image using the model pattern (a). Then, fill the area occluded by the head with pixels from the region connected to the front hair (c). Finally, perform image-to-image generation (d).}
    \vspace{-10pt}
    \label{fig:hair-repair}
\end{figure}

Once the final character image is established, the next step involves parsing the image and precisely positioning each component onto their respective layers. This process presents two main challenges. The first one is achieving pixel-level segmentation for each component, a task that remains difficult even when employing the SAM2 method \cite{ravi2024sam}. The second challenge involves addressing occluded areas. To address the first challenge, we use a template generated by the control mechanism as a mask to extract pixels directly from the original image. For the second challenge, we initially fill the occluded areas with pixels from the unoccluded regions, followed by the application of image-to-image control generation for refinement. For example, when restoring the back hair, which is largely obscured by the head (Fig.~\ref{fig:hair-repair}), we first erase the pixels of the head and then use image-to-image inpainting for reconstruction. This method not only prevents the generation of unwanted content but also ensures color consistency with the front hair.

\subsection{Animation}
\begin{figure}[tb]
    \centering
    \includegraphics[width=0.6\linewidth]{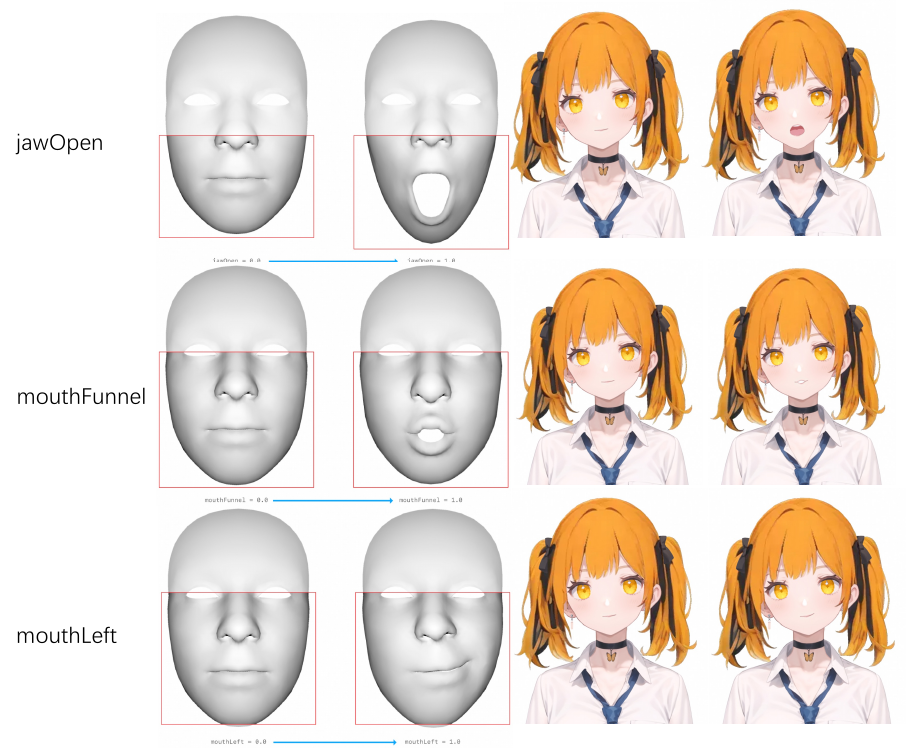}
    \caption{Live2D model supporting ARKit lip-sync driving.}
    \vspace{-10pt}
    \label{fig:arkit-live2d}
\end{figure}

\begin{figure}[tb]
    \centering
    \includegraphics[width=0.7\linewidth]{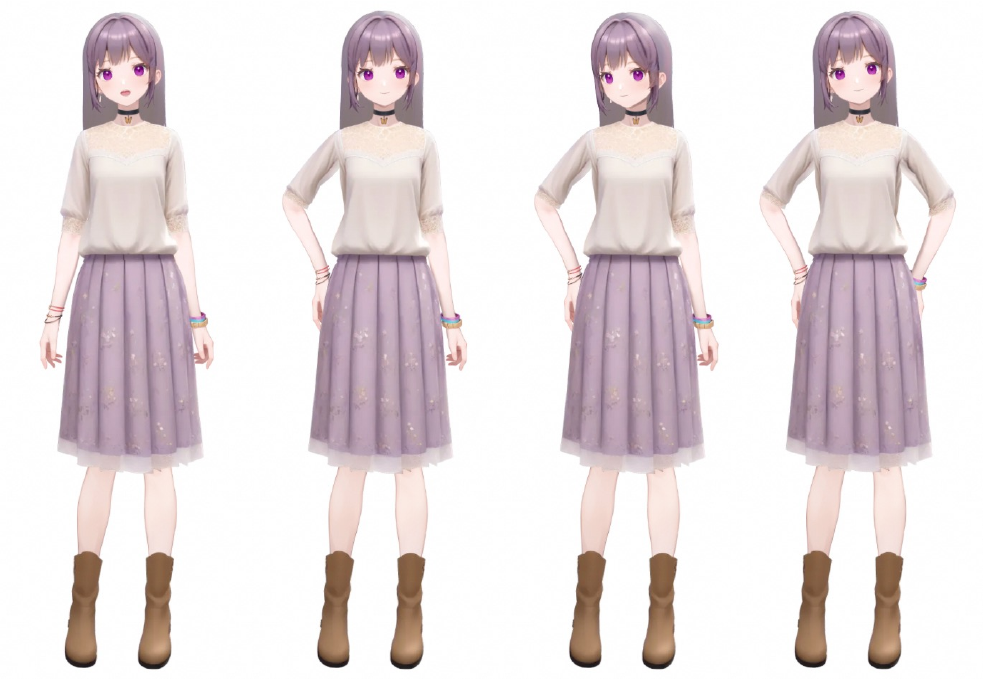}
    \caption{The overall animation effects of the generated Live2D model.}
    \vspace{-10pt}
    \label{fig:body-motion}
\end{figure}

Most Live2D models typically rely on just two parameters—MouthOpenY and MouthForm—for lip-syncing, leading to limited dynamic interactions. In contrast, ARKit \cite{arkitfaceblendshapes} offers 52 parameters to capture detailed facial expressions. Inspired by ARKit's extensive parameter set, we have developed more comprehensive lip-sync parameters, as illustrated in Fig.~\ref{fig:arkit-live2d}. By creating Live2D mouth blendshapes that correspond to the ARKit framework, we significantly enhance the liveliness and expressiveness of model animations, as demonstrated in Fig.~\ref{fig:body-motion}.

\begin{figure}[tb]
    \centering
    \includegraphics[width=1.0\linewidth]{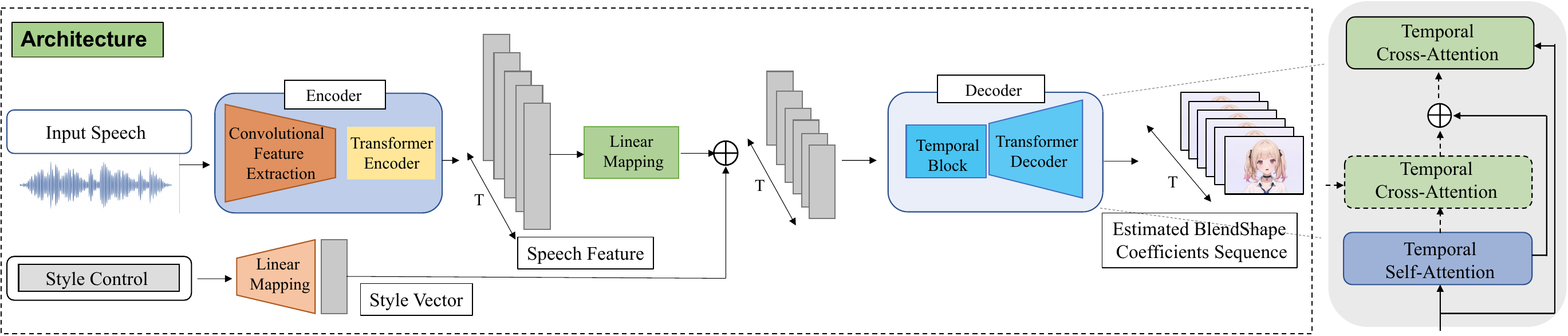}
    \caption{Animation framework. The Wav2Vec-XLSR~\cite{conneau2020wav2vec2,baevski2020wav2vec} is used to extract audio features from a speech signal. The audio features, combined with style features, are fed into a transformer to output stylized blendshape coefficients.}
    \vspace{-5pt}
    \label{fig:facial-animation-network}
\end{figure}

\begin{figure}[tb]
    \centering
    \includegraphics[width=1.0\linewidth]{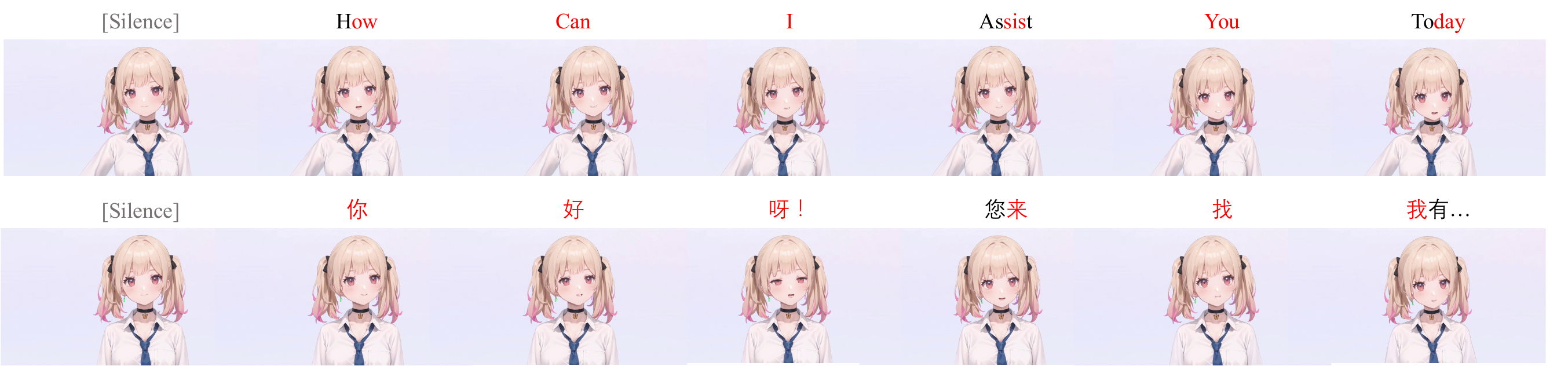}
    \caption{Visualization for audio-driven Live-2D face animation.}
    \vspace{-5pt}
    \label{fig:facial-animation-live2d}
\end{figure}

For the speech-driven facial animation, we designed a network that takes speech as input and outputs blendshape coefficients to achieve vivid facial expressions. As shown in Fig. \ref{fig:facial-animation-network}, the network follows an encoder-decoder architecture. We adopt the state-of-the-art pre-trained speech model Wav2Vec-XLSR~\cite{conneau2020wav2vec2,baevski2020wav2vec} for the audio encoder. The features extracted from raw audio waveform are combined with style features and fed into a transformer-based decoder, which outputs stylized blendshape coefficients. The Fig. \ref{fig:facial-animation-live2d} also illustrates the facial animation generated from English and Chinese speeches.

\section{Results}
\label{sec:Results}
Combining the modules mentioned above, our system is capable of generating a controllable, aesthetically pleasing, and drivable Live2D character based on a single sentence within one minute.  Fig.~\ref{fig:live2d-results} shows some of our generated outputs, which validate the effectiveness of our method in terms of visual appeal and diversity.

\section{Limitation}
\label{sec:Limitation}
Although we innovatively apply generative models to the creation of Live2D characters, enabling the automated generation of various 2D cartoon avatars, there are still some limitations. Firstly, our process relies on input text to generate Live2D characters, but text has difficulty conveying complex and nuanced information, making the controlled generation of details a challenge. Secondly, our generated results are constrained by the layout of the component layers in the original Live2D models, with a limited variety of component styles available.

\begin{figure*}[tb]
\centering
\includegraphics[width=0.98\textwidth]{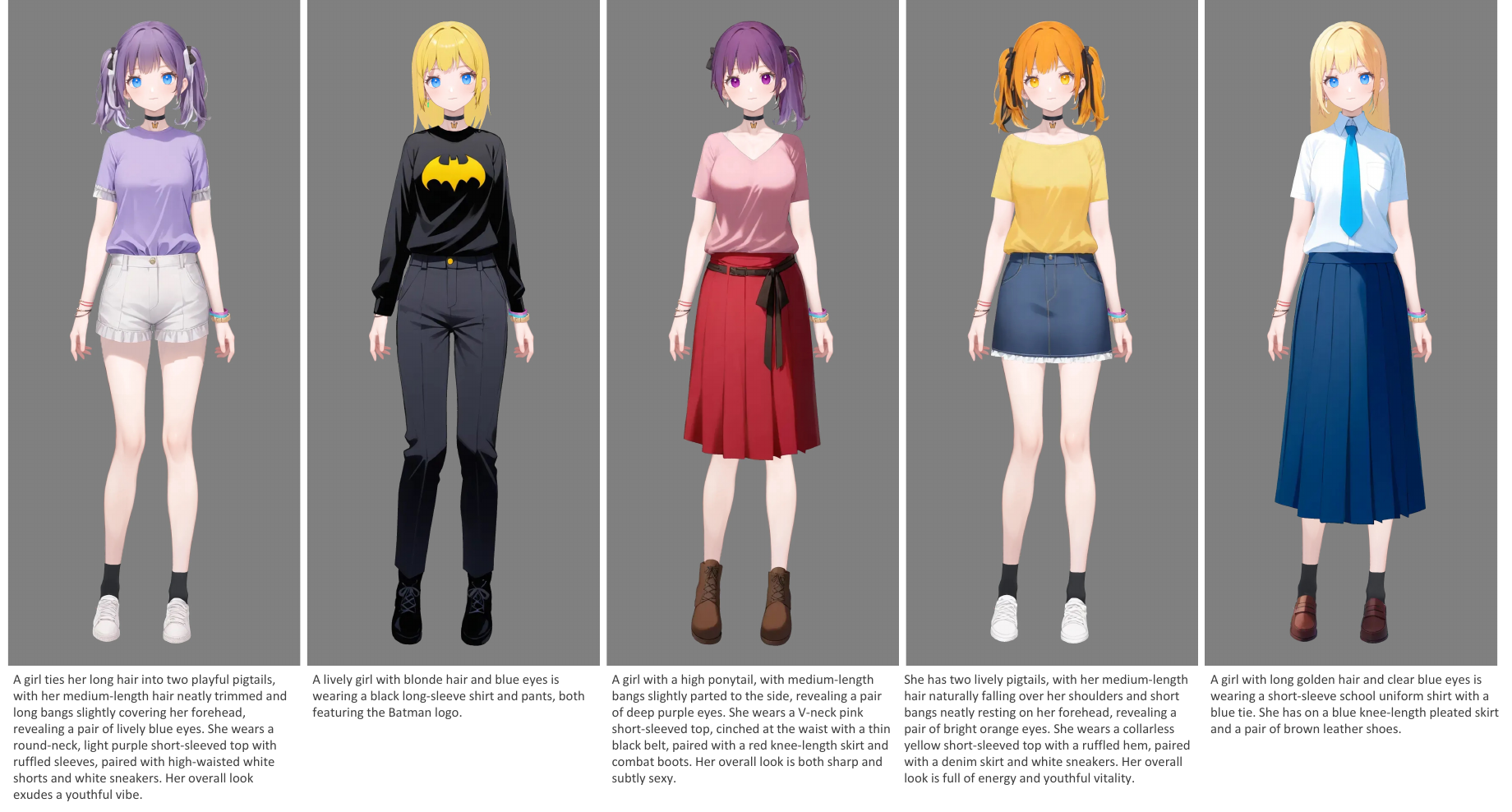}
\caption{Examples of Live2D cartoon characters created along with their corresponding text prompts.}
\label{fig:live2d-results}
\vspace{-15pt}
\end{figure*}

\section{Conclusion}
\label{sec:Conclusion}
 We present Textoon, the first method for generating diverse Live2D cartoon characters from text descriptions. By harnessing cutting-edge language and vision models, Textoon can quickly create a variety of stunning and interactive 2D characters in less than a minute. We also integrate ARKit-compatible facial blendshapes, enhancing mouth movements for more expressive interactions, allowing lively conversations with users. The live2D cartoon characters generated can be seamlessly rendered using HTML5, offering a wide range of application possibilities.

{\small
\bibliographystyle{ieeenat_fullname}
\bibliography{11_references}
}


\end{document}


\title{\paperTitle}
\author{\authorBlock}
\maketitlesupplementary

\appendix
\section{Appendix Section}
\label{sec:appendix_section}
Supplementary material goes here.

{\small
\bibliographystyle{ieeenat_fullname}
\bibliography{11_references}
}